\documentclass{article}

\usepackage{PRIMEarxiv}

\usepackage[utf8]{inputenc} 
\usepackage[T1]{fontenc}    
\usepackage{hyperref}       
\usepackage{url}            
\usepackage{booktabs}       
\usepackage{amsfonts}       
\usepackage{nicefrac}       
\usepackage{microtype}      
\usepackage{lipsum}
\usepackage{fancyhdr}       
\usepackage{graphicx}       
\graphicspath{{media/}}     

\title{Error Parity Fairness: Testing for Group Fairness in Regression Tasks
}

\author{
  Furkan Gursoy, Ioannis A. Kakadiaris \\
  Computational Biomedicine Lab, Dept. of Computer Science \\
  University of Houston \\
  Houston, TX, USA\\
  \texttt{\{fgursoy, ioannisk\}@uh.edu} \\
}

\begin{document}
\maketitle

\begin{abstract}
The applications of Artificial Intelligence (AI) surround decisions on increasingly many aspects of human lives. Society responds by imposing legal and social expectations for the accountability of such automated decision systems (ADSs). Fairness, a fundamental constituent of AI accountability, is concerned with just treatment of individuals and sensitive groups (e.g., based on sex, race). While many studies focus on fair learning and fairness testing for the classification tasks, the literature is rather limited on how to examine fairness in regression tasks. This work presents error parity as a regression fairness notion and introduces a testing methodology to assess group fairness based on a statistical hypothesis testing procedure. The error parity test checks whether prediction errors are distributed similarly across sensitive groups to determine if an ADS is fair. It is followed by a suitable permutation test to compare groups on several statistics to explore disparities and identify impacted groups. The usefulness and applicability of the proposed methodology are demonstrated via a case study on COVID-19 projections in the US at the county level, which revealed race-based differences in forecast errors. Overall, the proposed regression fairness testing methodology fills a gap in the fair machine learning literature and may serve as a part of larger accountability assessments and algorithm audits.

\end{abstract}

\section{Introduction}
Applications of artificial intelligence (AI) to decision-making tasks on humans have been becoming more common. Automated decision systems (ADSs) are widely employed in (i) finance \cite{caoAIFinanceChallenges2022} for credit scoring \cite{abdouCreditScoringStatistical2011}, real estate appraisal \cite{trawinskiComparisonExpertAlgorithms2017}, loan monitoring \cite{bozReassessmentMonitoringLoan2018}, and insurance premium estimation \cite{chapadosEstimatingCarInsurance2001}; (ii) healthcare \cite{MachineLearningHealthcare} for predicting disease risk based on genotype \cite{liElectronicHealthRecords2020}, forecasting COVID-19 spread in regional units \cite{zhengExploringInfluenceHuman2021}, and predicting high-cost high-need patients for inclusion in care management programs \cite{deruijterPredictionModelsFuture2022}; and (iii) other areas such as criminal justice and law enforcement \cite{zavrsnikCriminalJusticeArtificial2020}, predicting student performance \cite{namounPredictingStudentPerformance2021}, and border control \cite{borderArtificialIntelligenceEU}.

The increasing use of AI-based ADSs frequently becomes a target of criticism due to the biases such systems may exhibit. Apple and its banking partner Goldman Sachs became a target of popular criticism after Apple Card users reported lower limits for women \cite{telfordAppleCardAlgorithm2019}. Although an official investigation did not find proof of sex-based discrimination \cite{campbellAppleCardDoesn2021}, the case revealed the societal expectations and the risks organizations face if their products are shown or perceived to be biased. In other cases in the financial domain, gaps in home appraisal values are identified based on race and ethnicity \cite{narragonRacialEthnicValuation2021} and unexplained differences in insurance premiums are observed based on race \cite{angwinMinorityNeighborhoodsPay2017} as well as gender and birthplace \cite{fabrisAlgorithmicAuditItalian2021}. In the healthcare domain, genetic risk predictions are shown to not generalize well to different and understudied demographics \cite{martinHumanDemographicHistory2017}. COVID-19 spread forecast models produce dissimilar predictions hence caution is needed for public health strategies based on such models \cite{xiangCOVID19EpidemicPrediction2021}. Predicting healthcare costs for inclusion in care management programs leads to discrepancies in the treatments received by different racial groups \cite{obermeyerDissectingRacialBias2019}. Such problematic use cases of ADSs reach beyond the presented examples as bias is shown to exist in other real-world ADSs such as in criminal recidivism prediction \cite{angwinMachineBias2016}, facial recognition for law enforcement \cite{hillWrongfullyAccusedAlgorithm2020}, and welfare fraud detection \cite{vanbekkumDigitalWelfareFraud2021}.

The potentially problematic applications of AI in many aspects of human lives necessitate a critical lens toward AI. There are a growing number of studies and an expanding interest in critically examining ADSs for fairness, explainability, transparency, privacy, safety, and other societal expectations. This emerging field is known as AI Accountability, Responsible AI, Trustworthy AI, or AI Ethics. A fundamental problem in this field, fairness, is concerned with ADSs' potential unjust treatment of individuals or groups (e.g., based on sex, race, and other sensitive or protected characteristics).

Most group fairness notions require parity between chosen group-average statistics related to the decision outputs. For instance, for classification tasks, different fairness criteria require one (or a subset) of the statistics such as positive prediction ratio, precision, specificity, and recall to be equal across groups \cite{castelnovoClarificationNuancesFairness2022}. While the fairness literature on classification is relatively rich, the literature on fairness criteria for regression tasks is underdeveloped. This paper focuses on developing a fairness testing methodology for regression tasks. The introduced fairness notion, error parity, seeks the parity of prediction errors across sensitive groups. The proposed methodology can be incorporated into algorithm audits of ADSs for detecting fairness issues.

The main contributions of this paper are as follows.
\begin{enumerate}
    \item Error parity, a group fairness notion for regression tasks, is formally described and discussed.
    \item Error parity fairness is operationalized by designing appropriate statistical testing and inspection mechanisms.
        \begin{enumerate}
            \item An error parity test based on the comparison of the error distributions of different groups is presented to determine whether an ADS exhibits unfair behavior.
            \item A permutation test for comparing mean, variance, skewness, and kurtosis of the error distributions is presented as a post hoc analysis to explore the specific characteristics of any disparate treatment and to identify impacted groups.
        \end{enumerate}
\end{enumerate}

To facilitate fairness assessments of ADSs for regression tasks, a software program to assist with the analysis of error parity fairness is provided as an open-source tool\footnote{The software will be available upon publication.}. To illustrate the suitability and applicability of the proposed regression fairness testing methodology, a demonstrative case study on the fairness of the US county-level COVID-19 projections is conducted using the provided software.

\section{Related Work}\label{sec:rl}
The emerging interdisciplinary field that is loosely referred to as AI Accountability, Responsible AI, Trustworthy AI, or AI Ethics has attracted substantially increasing research interest and funding. The fields' own conferences such as FAccT\footnote{\noindent ACM Conference on Fairness, Accountability, and Transparency (https://facctconference.org/, https://www.fatml.org/) began as a workshop in 2014.}, 
AIES\footnote{AAAI/ACM Conference on AI, Ethics, and Society (https://www.aies-conference.com/) began in 2018.}, 
EAAMO\footnote{ACM conference on Equity and Access in Algorithms, Mechanisms, and Optimization (https://eaamo.org/) began in 2021.}, and
SATML\footnote{IEEE Conference on Secure and Trustworthy Machine Learning (https://satml.org/) will begin in 2023.} emerged in recent years in addition to recently founded journals and conference special tracks allocated to the field. AI Accountability encompasses problems in AI that are centered around topics such as transparency, documentation, fairness, explainability, privacy, security, safety, risks, assurance, and applied ethics. Several studies \cite{koshiyamaAlgorithmAuditingSurvey2021, gursoySystemCardsAIbased2022, kaurTrustworthyArtificialIntelligence2022} provide relatively comprehensive overviews of the field.

Fairness, used as a relatively vague concept in the broader machine learning literature, is concerned with tackling the unjust treatment of individuals or groups.  Individual fairness is interested in whether similar individuals receive similar treatments whereas group fairness is interested in the disparities in the average treatments of sensitive groups. This paper focuses on group fairness for which various approaches and notions exist. The two principles governing different fairness approaches are equal outcome and equal opportunity. Equal outcome (also referred to as statistical or demographic parity) stipulates that the outputs of a decision system (e.g., predictions from a machine learning model) should have the same distribution across all groups. In other words, sensitive variables and the outputs should be independent. Equal opportunity, on the other hand, is satisfied when sensitive variables and the outputs are conditionally independent given the underlying merit, e.g., given ground truth values. Therefore, equal opportunity is more concerned with the prediction errors whereas equal outcome is concerned with the predictions regardless of the ground truth values.

In the machine learning fairness literature, the overwhelming majority of the methods, tests, metrics, and other analyses are primarily or exclusively designed for and applicable to classification tasks where ADS outputs take values from a finite set of class labels. Another group focuses on ranking problems where ADS outputs are an ordered list of the input instances. In contrast to the classification and ranking tasks, regression tasks are aimed at predicting continuous variables. Compared to the fair classification and fair ranking literature, the fair regression literature is limited. Interested readers may refer to the comprehensive surveys of the fairness literature in general \cite{mehrabiSurveyBiasFairness2021} as well as the reviews focusing on fairness for classification \cite{pessachReviewFairnessMachine2022} and fairness for ranking \cite{zehlikeFairnessRankingPartI2022, zehlikeFairnessRankingPartII2022}. In addition, Fabris et al. \cite{fabrisAlgorithmicFairnessDatasets2022} provide a survey of datasets for fairness analysis. A brief overview of the regression fairness literature is provided next to highlight the novelty of the fairness testing methodology proposed in this paper.

A large majority of the fair regression literature is concerned with designing fair learning methods rather than developing testing or measuring mechanisms for fairness.
Mohamed and Schuller \cite{mohamedNormaliseFairnessSimple2022} propose methods to preprocess the training data for machine learning models so that the sample size and the target variable distribution are on par across sensitive groups. Berk et al. \cite{berkConvexFrameworkFair2017} and Pérez-Suay et al. \cite{perez-suayFairKernelLearning2017} propose fairness regularizers, i.e., specifically designed penalty terms included in the loss function of the learning procedure so that the optimization is softly constrained to learn fair predictors.
Fukuchi, Sakuma, and
Kamishima \cite{fukuchiPredictionModelBasedNeutrality2013} model sensitive group memberships as probabilities and perform maximum likelihood estimation to remove dependencies between sensitive variables and predictions. Chzhen et al. \cite{chzhenFairRegressionWasserstein2020} propose a method to update the predictions in a post hoc manner such that fairness constraints are satisfied.

There are also a few studies that provide relatively tangible fairness criteria for regression tasks. Komiyama et al. \cite{komiyamaNonconvexOptimizationRegression2018} define fairness as the proportion of variance in the target variable that is explained by sensitive variables. Their criterion seeks the independence of sensitive variables and predictions, hence following the equal outcome principle only. Calders et al. \cite{caldersControllingAttributeEffect2013} propose two separate fairness criteria based on the equal outcome and equal opportunity principles. The first criterion requires the parity of the mean predictions across groups whereas the second requires the parity of mean residuals (i.e., differences between the predictions and the ground truth values). This method compares the distributions only by their means without accounting for potential disparities in the shapes of the distributions and without any statistical significance testing. 

Agarwal, Dudik, and Wu \cite{agarwalFairRegressionQuantitative2019} employ a more holistic approach in comparing distributions of predictions and residuals where the Kolmogorov-Smirnov (KS) distance \cite{masseyKolmogorovSmirnovTestGoodness1951} is computed to quantify the differences between group distributions and the overall distribution in their fair learning method. However, they do not use the KS distance for testing hypotheses on whether the group distributions come from the same population. Chzhen et al. \cite{chzhenFairRegressionWasserstein2020} also employ the KS distance to quantify differences in the prediction distributions (i.e., following the equal outcome principle) without any statistical hypothesis testing. Perera
et al. \cite{pereraSearchbasedFairnessTesting2022} propose a fairness measure based on a worst-case scenario. Specifically, they propose a searching algorithm to identify the maximum difference in predictions when the only change in the input is the sensitive group membership. However, such a measure is easily dominated by rare extreme cases and cannot reliably reflect the overall group-based disparities. 

In conclusion, while a considerable fair learning literature exists for regression, there is a gap for statistically sound and comprehensive testing mechanisms to audit ADSs for fairness in numeric prediction tasks. The fairness testing methodology presented in the next section differs from the existing regression fairness criteria in two ways. First, it is bolstered by appropriate robust statistical hypothesis testing. Second, it provides a comprehensive post hoc analysis to identify and interpret the specific characteristics of potential fairness issues. Therefore, the proposed methodology associated with the error parity fairness notion is reliable and actionable in real-world algorithm audits of AI systems.

\section{Error Parity Fairness}\label{sec:rpf}

\subsection{Notation}
Scalar values are denoted by lower case letters (e.g., $a$).
Vectors are denoted by boldface lowercase letters (e.g., $\mathbf{a}$). The $i^{th}$ element of $\mathbf{a}$ is denoted as $\mathbf{a}_i$.
A vector filtered by a criterion $i$ is denoted as $\mathbf{a}^i$.
Matrices are denoted by boldface uppercase letters (e.g., $\mathbf{A}$). 
The real value space and the categorical value space are denoted respectively by $\mathbb{R}$ and $\mathbb{S}$.
A vector of categorical values with size $n$ is denoted as  $\mathbf{a} \in \mathbb{S}^{n}$. A function is denoted as $f: \bullet \rightarrow \bullet$.

\subsection{Problem Definition}

This paper proposes an error parity test and a post hoc permutation test to investigate the fairness of a decision system, given the following:

\begin{itemize}

    \item a matrix $\mathbf{X}$ that represents $n$ observations and $m$ features where $\mathbf{X} \in \mathbb{(R \cup S)}^{n \times m}$,
    
    \item a vector $\mathbf{s}$ that represents the sensitive group memberships for the $n$ observations where $\mathbf{s} \in \mathbb{S}^{n}$  regardless of whether $\mathbf{s} \perp \mathbf{X}$ in general,

    \item a decision system $f: \mathbf{X} \rightarrow \mathbf{\hat{y}}$,
    
    \item decision outputs $\mathbf{\hat{y}}$ where $\mathbf{\hat{y}} \in \mathbb{R}^{n}$, and
    
    \item corresponding ground truth values $\mathbf{y}$ where $\mathbf{y} \in \mathbb{R}^{n}$.
    
\end{itemize}

The proposed error parity fairness notion stipulates that the prediction errors must have the same distribution for all sensitive groups. To this end, first, a statistical test examining whether different data come from the same distribution is employed to determine whether a decision system is fair or not. Second, a permutation test is presented to assess any group-based statistically significant differences in the first four moments of the error data (i.e., mean, variance, skewness, and kurtosis) to inspect the disparities. 

\subsection{Error Metric Selection}

To determine whether a decision system is fair or not, error parity fairness investigates the relationship between sensitive groups and errors. Let $e: (\mathbf{y}, \mathbf{\hat{y}}) \rightarrow \mathbf{r}$ to represent the errors in general. There may be different choices for the function $e$ depending on the specific problem, domain, and application. Several candidates are presented below.

\begin{itemize}
    \item \textit{Differences.} The most simple approach is to compute the difference vector between the predicted and actual values, i.e., $\mathbf{r} = \mathbf{\hat{y}} - \mathbf{y}$. This approach is suitable when the non-relative magnitude of deviations is of primary concern and the direction of the deviations is important.
    \item \textit{Absolute differences.} When the direction of the deviations does not matter, absolute differences ($\mathbf{r} = |\mathbf{\hat{y}} - \mathbf{y}|$) may also be employed. This choice would be rarely relevant in the real world since underprediction and overprediction have different implications in most cases.
    \item \textit{Squared differences.} When larger deviations require an exponentially increasing penalty, a squared differences approach may be employed. Depending on whether the direction of the deviation is important, the original signs may be retained (i.e., $\mathbf{r} = (\mathbf{\hat{y}} - \mathbf{y}) |(\mathbf{\hat{y}} - \mathbf{y})|$) or may be dismissed (i.e., $\mathbf{r} = (\mathbf{\hat{y}} - \mathbf{y})^2$).
    In this case, outlier errors have more impact on the test results.
    
    \item \textit{Percentage-based differences.} Often, the relative deviations are more relevant than their absolute values. The same amount of deviation may be relatively small or large depending on the magnitude of the ground truth value. Then, a simple choice would be the percentage error based on ground truth values, i.e., $\mathbf{r} = (\mathbf{\hat{y}} - \mathbf{y})/\mathbf{y}$. However, if ground truth values are too small, such percentage errors may be inflated. If the ground truth is zero, the percentage error becomes intractable. Alternatively, the errors can be computed based on the average of the predicted and ground truth values, i.e., \mbox{$\mathbf{r} = (\mathbf{\hat{y}} - \mathbf{y})/ (\mathbf{y} + \mathbf{y})$}. If both ground truth and prediction error values are zero, this function would be indeterminate and should be externally set to zero.
\end{itemize}

\subsection{Error Parity Test}

In general terms, error parity fairness requires that $\mathbf{r} \perp \mathbf{s}$. Given that $\mathbf{r}^i$ is the subset of $\mathbf{r}$ that corresponds to the group $i$, if such independence indeed exists, it follows that all $\mathbf{r}^i$ come from the same population distribution. There are various statistical tests in the literature for comparing distributions as a whole. A popular choice is the Kolmogorov-Smirnov (KS) test \cite{masseyKolmogorovSmirnovTestGoodness1951, bergerKolmogorovSmirnovTest2014}.
While it is possible to extend it multiple-sample comparisons \cite{conoverSeveralKSampleKolmogorovSmirnov1965}, its \textit{k}-sample implementation is not widely available. Moreover, the KS test has low sensitivity for the differences in the tails of the distribution \cite{goldmanComparingDistributionsMultiple2018}. Alternatively, the Anderson-Darling (AD) test \cite{andersonAsymptoticTheoryCertain1952} has a widely available \textit{k}-sample implementation for comparing multiple distributions and better sensitivity to the tails of the distributions \cite{goldmanComparingDistributionsMultiple2018}. Engmann and Cousineau \cite{engmannCOMPARINGDISTRIBUTIONSTWOSAMPLE} argue that the AD test, the KS test, and multiple other alternatives have similar underlying structures, the AD test is more powerful and requires less data than the KS test, and ultimately the AD test is a superior choice in general. Thus, this study employs the AD test for investigating whether the observed differences in group-based error distributions are statistically significant. The formal description and the implementation details of the \textit{k}-sample AD test are provided by Scholz and Stephens \cite{scholzKSampleAndersonDarlingTests1987} and Virtanen et al. \cite{2020SciPy-NMeth}.

If the AD test fails to reject the null hypothesis that the distributions come from the same population, it is determined that the decision system is fair. Otherwise, it is determined that there is at least one group whose distribution does not come from the same population. However, this does not reveal which distributions are different or what are the characteristics of such differences (e.g., differences in mean, variance, skewness, kurtosis). Therefore, a subsequent analysis is needed. To this end, this paper employs a permutation test-based approach as a post hoc analysis to follow a statistically significant result obtained from the AD test. 

\subsection{Permutation Tests}

The shape of a distribution may be described by its moments in general.
The first raw moment corresponds to the mean ($\mu$), the second central moment corresponds to the variance ($\sigma^2$), and the third and fourth standardized moments correspond to the skewness ($\psi$) and the kurtosis ($\kappa$) respectively. The next moments are called high-order moments and they become more difficult to estimate and interpret as the order increases, therefore, are excluded from this paper. The first four moments are mathematically defined in Equations 1 through 4 where $\mathrm{E}\left[\bullet\right]$ is the expectation function and $x$ is the random variable representing the distribution.

\begin{equation}
    \mu = \mathrm{E}\left[x\right]
\end{equation}

\begin{equation}
    \sigma^2 = \mathrm{E}\left[\left({x-\mu}\right)^{2}\right]
\end{equation}

\begin{equation}
    \psi = \mathrm{E}\left[\left(\frac{x-\mu}{\sigma}\right)^{3}\right]
\end{equation}

\begin{equation}
    \kappa = \mathrm{E}\left[\left(\frac{x-\mu}{\sigma}\right)^{4}\right]
\end{equation}

While parametric and non-parametric statistical tests exist for comparing means and variances of different samples, no such tests exist for comparing skewness and kurtosis, to the best of the authors' knowledge. Moreover, most statistical tests have underlying assumptions that are often strong. This is particularly true and more strict for parametric tests. 

For testing the equality of means, the one-way ANOVA test assumes that the distributions are normal and have equal variances \cite{sawyerAnalysisVarianceFundamental2009}. Neither of these assumptions can be safely made in the case of error distributions from arbitrary machine learning-based ADSs. The popular non-parametric alternative, the Kruskal-Wallis test \cite{kruskalUseRanksOneCriterion1952}, does not require normality. However, it assumes that the distributions are identically shaped and scaled \cite{ostertagovaMethodologyApplicationKruskalWallis2014}. In the case of the fairness analysis required here, this cannot be safely assumed either. In connection, the Behrens-Fisher problem is concerned with the hypothesis testing for the means when the variances are not equal. The problem still remains to be an area of discussion \cite{kimBehrensFisherProblemReview1998}. The Welch's test proposes a solution but maintains the normality assumption \cite{welchGENERALIZATIONSTUDENTPROBLEM1947}. Consequently, to the best of the authors' knowledge, there is no statistical test for comparing multiple means whose assumptions are acceptable for the proposed fairness testing procedure.

For testing the equality of variance, the F-test of equality of variance for two groups and its generalizations to multiple groups, Hartley's test \cite{hartleyMAXIMUMFRATIOSHORTCUT1950} and Bartlett's test \cite{bartlettPropertiesSufficiencyStatistical1937} assume that the values for all groups are normally distributed. Alternatively, some tests such as Levene's test, O'Brien test, and the Brown-Forsythe test are more robust to non-normality \cite{brownRobustTestsEquality1974, alginaTypeErrorRates1989}. Algina, Olejnik,
and Ocanto \cite{alginaTypeErrorRates1989} found that O'Brien and Brown-Forsythe tests have no consistent advantage over another and each may perform better for different distributions.

For testing the equality of skewness, there is at least one method \cite{randlesAsymptoticallyDistributionFreeTest1980} for testing the symmetry of a single sample (i.e., testing for statistically significant skewness). However, to the best of the authors' knowledge, no test exists for examining the equality of skewness for two or more samples. Similarly, to the best of the authors' knowledge, no statistical test exists for examining the equality of kurtosis for two or more samples.

Given the strong assumptions of the tests for comparing means and the lack of tests for comparing skewness and kurtosis, this study resorts to permutation tests. A permutation test approach is preferred for three reasons. First, permutation tests only assume exchangeability of the labels (e.g., group memberships) \cite{lafleurIntroductionPermutationResamplingBased2009}. Second, the same permutation test approach can be used for comparing all four statistics (and any additional statistics) hence providing a single unified framework that can incorporate all potentially relevant measures. Third, permutation tests are exact \cite{lafleurIntroductionPermutationResamplingBased2009} and sampling-based permutation tests are also exact when the sample size is large \cite{lafleurIntroductionPermutationResamplingBased2009, manlyRandomizationBootstrapMonte1997}.

The proposed permutation test works as follows. First, the group memberships are randomly shuffled $l$ times and the four statistics (mean, variance, skewness, and kurtosis) are computed each time. Then, the group differences in the simulated cases and the observed case are compared. If the absolute differences in the original observation are larger than the corresponding differences in fraction $\alpha$ of the $l$ simulated cases, \textit{p-value} becomes $1-\alpha$. While the test may easily become computationally prohibitive if all permutations are considered, a randomized version with sufficiently large value of $l$ (typically $>5000$ for the \textit{p-value} of $0.01$ \cite{marozziRemarksNumberPermutations2004}) is computationally feasible.

Following a statistically significant result from the error parity test, the permutation test is to be conducted for all sensitive group pairs. For any statistically significant difference identified via the permutation test, the observed values can be compared and interpreted. For example, consider an ADS that outputs personal credit card limits for a set of customers, the sensitive groups based on sex ($f$: female, $m$: male) and the error metric $\mathbf{r} = \mathbf{\hat{y}} - \mathbf{y}$. Following are sample interpretations of potential differences in each statistic.

\begin{itemize}

    \item The interpretation of a difference in means is straightforward. The group with the larger mean is overpredicted on average. For instance, if $\mu_{m} > 0 > \mu_{f}$, it follows that males are in an advantageous position since they receive higher limits while females are underserved in terms of credit limits.
    
    \item The group with the larger variance has more dispersed prediction errors on average. For instance, if $\sigma^2_{m} > \sigma^2_{f}$, assuming that $\mu_{m} = \mu_{f}$, it follows that females receive more consistent predictions whereas some males benefit more from overprediction at the cost of other males suffering more from underprediction.
    
    \item Positive and negative skewness indicates a longer tail on the right and the left side of the distribution, respectively. Zero skewness indicates symmetry.
    It follows that the group with a positive larger skewness value has more members who are substantially overpredicted, resulting in an advantageous position. However, if the group means are equal, such an advantage is at the cost of other members of the same group who are relatively underpredicted on average. For instance, assuming equality for means and variances, if $\psi_{m} > 0 > \psi_{f}$, it indicates that few males (females) are in a highly advantageous (disadvantageous) position at the cost (benefit) of the other members of their groups. If $\psi_{m} > 0 = \psi_{f}$ or $\psi_{m} < 0 = \psi_{f}$, it may also be argued that the model does not learn well for males since the error distribution is not symmetric.
    
    \item Using Fisher's definition of kurtosis, the zero value corresponds to the kurtosis value of, for instance, a normal distribution. Positive kurtosis indicates fat tails whereas negative kurtosis indicates thin tails. Therefore, a group with a higher kurtosis has more outliers  \cite{westfallKurtosisPeakedness19052014}. Assuming parity in other statistics, if $\kappa_{m} > \kappa_{f}$, there are more males than females that are extremely underpredicted and/or overpredicted. Similar to skewness, kurtosis alone does not usually allow to conclude whether one group is more advantageous than the other. However, it suggests that the model does not learn very well for groups with higher kurtosis as the corresponding errors contain more outliers.
    
\end{itemize}

\subsection{Scope, Discussion, and Limitations}

The scope and applicability of the proposed regression fairness testing methodology are discussed next.

\begin{itemize}
    \item The proposed error parity fairness is primarily designed for regression tasks. However, it may be applied
    to the binary classification tasks where predicted probabilities can be obtained from the ADS.
    
    \item The error parity fairness conforms to the equal opportunity principle. However, the presented methodology can be easily modified to adapt to equal outcomes. By setting $\mathbf{r} = \mathbf{\hat{y}}$, hence ignoring ground truth values and comparing predictions rather than errors, the proposed methodology tests for the equal outcomes principle.
    
    \item A representative and acceptable test set is required. Moreover, ground truth values must be available, reliable, and unbiased. In practice, such qualities for a test set cannot be determined with absolute certainty. Therefore, a continuous fairness monitoring approach is highly recommended such that the ADS is assessed based on the real-world data it operates on during its actual use. 
    
    \item The minimum sample size required for possibly obtaining a statistically significant result is small. In the case of two groups, given that $a$ and $b$ are the sizes of the two groups where $n = a + b$, there are $n! / a!b!$ unique permutations. To obtain the statistical significance level of $0.01$, the number of unique permutations must be at least 200. The observed case and the case where the labels are swapped would constitute the two extreme cases where no other combination can produce a larger difference, hence $2/200 = 0.01$. When the group sizes are equal, the minimum sample size becomes $n=10$. Consequently, the proposed methodology works well even with small sample sizes. Moreover, as ADSs typically work with a much larger number of observations, the sample size would not be an area of concern in typical cases.
    
    \item  The proposed error parity notion does not prescribe a specific function for computing the errors. The developers and auditors of an ADS may choose a function based on appropriate error metrics in the literature or devise a custom function as long as it is real-valued and reasonably supported by the nature of the decision problem. If there is more than one plausible alternative, the analysis should be repeated and reported for each.  
    
    \item If there are multiple types of sensitive groups (e.g., both sex and race), the test should be repeated for each. Moreover, with an intersectional lens, it is also recommended to conduct the analysis with intersectional groups (e.g., Asian males, Asian females, Caucasian males, Caucasian females, and so on).
    
    \item If there is more than one dependent variable, the test can be repeated for each dependent variable separately. Alternatively, the test can be conducted on meaningful combinations of the dependent variables. For instance, if a decision system produces credit limits for different credit products (e.g., home, car, personal loan), it is appropriate to run a fairness analysis for the sum of all credit limits.

\end{itemize}

\section{Case Study}\label{sec:case}

The relationship between race and ethnicity and access to health and healthcare is a highly relevant research area \cite{richardsonAccessHealthHealth2010, webbhooperCOVID19RacialEthnic2020}. Healthcare policies, especially those relating to public health, may be informed by relevant forecasts. For instance, vaccine production and distribution may be prioritized based on the forecasted demand in space and time. Other mobile healthcare resources such as traveling nurses may also be stationed based on such projections. Consequently, the accuracy and fairness of ADSs that produce such forecasts need to be ensured.

A recent and popular public health-related forecast problem is to predict the number of new COVID-19 cases. Especially when a novel condition such as COVID-19 is spreading wide and fast, policy and intervention decisions need to be made faster. If the forecasts are not demographically fair, the policies informed by those forecasts may lead to inequitable access to health and healthcare. Accordingly, this brief case study demonstrates the need for and the application of the proposed fairness testing assessment on forecasts of new COVID-19 cases across the US counties.

\subsection{Methodology}

The data are collected from multiple sources. The COVID-19 new case predictions are obtained from a US federal agency program that brings forecasts from its partners and also provides an ensemble model \cite{cdcCoronavirusDisease20192020}. The numbers on actual cases are obtained from a data curation program that monitors official reports from states, counties, and regional health departments \cite{timesCoronavirusDataFrequently2020}. The data on racial demographics are obtained from the Annual Resident Population Estimates of the US Census Bureau \cite{bureauPopulationHousingUnit}. The time frame of the analysis is limited to the year 2021.

The numbers of new actual cases are computed based on the weekly changes in the cumulative numbers. However, sometimes governments revise the reported numbers in an ambiguous manner, resulting in negative values when computing new cases. Moreover, the populations of some counties are very small and their inclusion may skew the overall analysis. For these reasons, the week-county pairs with less than 100 new cases are removed.

Predictions are available from more than 20 partners (i.e., models). However, these predictions are not consistently available across all weeks and all counties. There are 3,142 counties and equivalent entities in the 50 states of the US, resulting in 163,384 prediction points for the ensemble model. The models with less than 150,000 prediction points are removed from the analysis. The remaining models come from Johns Hopkins University Applied Physics Lab (JHU), Facebook (FB), Columbia University (CU), and the University of Virginia (UVA) \cite{cdcCoronavirusDisease20192020}.

The demographic data contains many categorizations based on race and ethnicity. For simplicity, we compare two groups: The non-Hispanic White-alone (NHWA) population versus others. Slightly less than 60\% of the national population belongs to the NHWA group. Accordingly, 2,494 counties whose NHWA population proportions are greater than the national proportion are labeled as White (W), and 648 other counties are labeled as Non-White (N).

The described preprocessing results in a final dataset of five models, each with at least 44,000 prediction points for weekly new COVID-19 cases during 2021 across US counties and the corresponding actual new case numbers. The 76\% of the prediction points are for the counties labeled as W. Next, the proposed fairness testing methodology is employed to test for error parity fairness and investigate specific disparities in the error distribution. As an error metric, percentage error (i.e., $\mathbf{r} = (\mathbf{\hat{y}} - \mathbf{y})/\mathbf{y}$)  is employed since the county populations are substantially different. For the post hoc test, the number of random permutations is set to 100,000, an order of magnitude larger than the recommended minimum in the literature. Since a very large number is employed, the findings do not vary in different runs.

\subsection{Findings}

The mean (standard deviation) of the absolute percentage errors is $31\%$ ($29\%$) for the CU model, $41\%$ ($26\%$) for the FB model, $36\%$ ($66\%$) for the JHU model, $35\%$ ($45\%$) for the UVA model, and $26\%$ ($23\%$) for the ensemble model. The individual models, with the potential exception of the CU model, either have high errors (i.e., low accuracy) or high standard deviation (i.e., low consistency). The ensemble performs relatively well in terms of accuracy and consistency.

For all models, at the statistical significance level of $0.01$, the AD tests rejected the null hypothesis that the error distributions for different groups come from the same population distribution. Consequently, it can be concluded that the forecast errors have disparate behavior based on the racial composition of the counties. As models are unfair according to our analysis, the AD test is followed by the post hoc analysis to inspect the disparities in the error behavior. Table \ref{tab:posthoc} summarizes its results.

\begin{table}[t]
\centering
\begin{tabular}{r|rr|rr|rr|rr|}
\cline{2-9}
\multicolumn{1}{l|}{}      & \multicolumn{2}{c|}{$\mu$ (\%)}                            & \multicolumn{2}{c|}{$\sigma$ (\%)}                        & \multicolumn{2}{c|}{$\psi$}                              & \multicolumn{2}{c|}{$\kappa$}                      \\ \cline{2-9} 
\multicolumn{1}{l|}{}      & \multicolumn{1}{c|}{W}            & \multicolumn{1}{c|}{N} & \multicolumn{1}{c|}{W}           & \multicolumn{1}{c|}{N} & \multicolumn{1}{c|}{W}          & \multicolumn{1}{c|}{N} & \multicolumn{1}{c|}{W}   & \multicolumn{1}{c|}{N}  \\ \hline
\multicolumn{1}{|l|}{CU} & \multicolumn{1}{r|}{\textbf{-9}}  & \textbf{-6}            & \multicolumn{1}{r|}{\textbf{40}} & \textbf{46}            & \multicolumn{1}{r|}{3}          & 2                      & \multicolumn{1}{r|}{86}  & 25                      \\ \hline
\multicolumn{1}{|l|}{FB}   & \multicolumn{1}{r|}{\textbf{-30}} & \textbf{-16}           & \multicolumn{1}{r|}{\textbf{39}} & \textbf{46}            & \multicolumn{1}{r|}{1}          & 2                      & \multicolumn{1}{r|}{4}   & 11                      \\ \hline
\multicolumn{1}{|l|}{JHU}   & \multicolumn{1}{r|}{\textbf{0}}   & \textbf{6}             & \multicolumn{1}{r|}{\textbf{59}} & \textbf{112}           & \multicolumn{1}{r|}{17}         & 31                     & \multicolumn{1}{r|}{940} & 1550                    \\ \hline
\multicolumn{1}{|l|}{UVA}  & \multicolumn{1}{r|}{\textbf{0}}   & \textbf{3}             & \multicolumn{1}{r|}{55}          & 66                     & \multicolumn{1}{r|}{6}          & 9                      & \multicolumn{1}{r|}{122} & 286                     \\ \hline
\multicolumn{1}{|l|}{Ens.} & \multicolumn{1}{r|}{\textbf{-10}} & \textbf{-6}            & \multicolumn{1}{r|}{\textbf{33}} & \textbf{36}            & \multicolumn{1}{r|}{\textbf{1}} & \textbf{2}             & \multicolumn{1}{r|}{6}   & \multicolumn{1}{r|}{13} \\ \hline
\end{tabular}

\vspace{8pt}
\caption{The mean $\mu$, standard deviation ($\sigma$), skewness ($\psi$), and kurtosis ($\kappa$) values of the errors by forecast models for counties labeled as W and N. Instead of $\sigma^2$, $\sigma$ is reported  to have the same unit ($\%$) as the mean. Skewness and kurtosis are unitless. Values are displayed in bold when the associated W-N difference for the corresponding statistic is found to be statistically significant at the $0.01$ level.
When predictions for counties labeled as W are compared to counties labeled as N, the mean errors are significantly lower in all models, the error variance is significantly lower except for the UVA model, and the skewness value is significantly lower for the ensemble (Ens.) model. No statistically significant difference is observed for the kurtosis. 
} 
\label{tab:posthoc}
\end{table}

\begin{figure}[t]
\centering
\includegraphics[width=0.6\columnwidth]{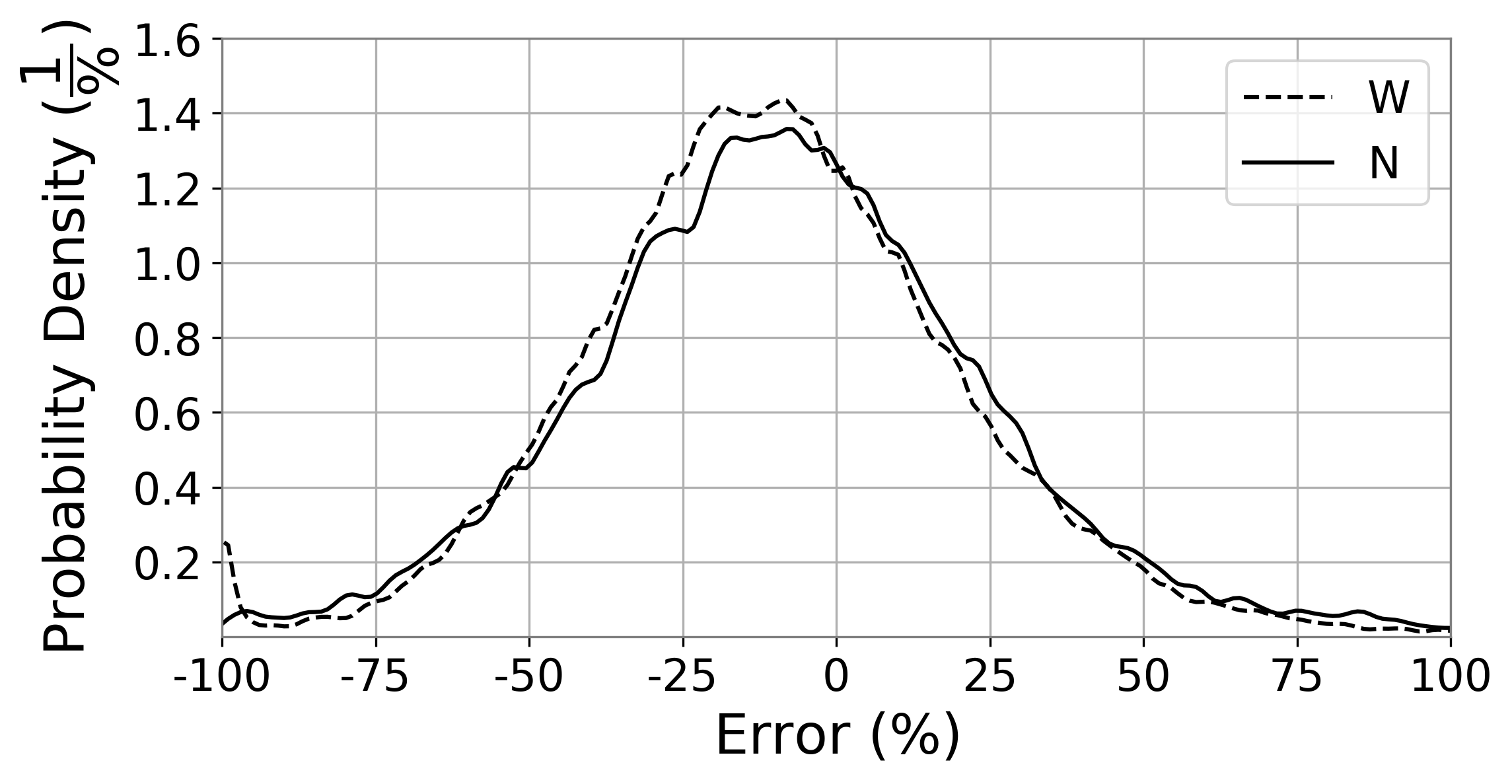}
\caption{Distribution of the errors from the ensemble model for counties labeled as W and N. The x-axis is cut at the value 100\%, i.e., the predicted values greater than double the actual values are not included. The excluded cases constitute 0.40\% and 0.99\%  of the predictions for counties labeled as W and N, respectively. The mean of the excluded error values is 149\% for W and 156\% for N. Although the two lines closely follow each other, the dashed line is generally above the solid line when the error is negative and the opposite is true when the error is positive. Moreover, for the dashed line, relatively many errors are concentrated around -100\% error, i.e., predicted values of zero. This visual inspection is in line with the earlier finding that COVID-19 risks are underpredicted for counties labeled as W but also hints that a considerable portion of this result is actually due to predictions concentrated on zero.}
\label{fig:dist}
\end{figure}

The lower (and negative) mean percentage error for the counties labeled as W indicates that COVID-19 spread risks are relatively underestimated for those counties. While the exact implications are context-specific, it may lead to fewer healthcare resources being allocated to them. On the other hand, it may also lead to less strict COVID-19 restrictions. The higher variance for counties labeled as N implies that the COVID-19 predictions are less consistent, which may lead to inadequate healthcare policies in those counties. However, it should be noted that, for most models, although statistically significant, the difference is rather small and may be inconsequential. The prediction errors from the ensemble model are more skewed for counties labeled as N, suggesting that the model does not learn well for those counties. Although large differences are observed in kurtosis for some models, none are found as statistically significant at the 0.01 or even at the 0.05 level. It indicates that the observed differences cannot be reliably taken at face value without proper statistical hypothesis testing. The visual inspection provided in Figure \ref{fig:dist} confirms these findings and provides a complementary analysis. 

\subsection{Remarks}

This brief demonstrative case study shows the usefulness of the proposed regression fairness testing methodology. It finds that the COVID-19 projections published by the leading public health institute in the US may be demographically biased. Specifically, the COVID-19 risks for counties labeled as W are relatively underpredicted on average whereas the predicted risks for counties labeled as N are less consistent on average. The potential implications of such disparate forecasting behavior are best left to the health policy experts.

\textit{Limitations:} It should be noted that the presented case study is limited in scope and in detail. First, although a very comprehensive dataset, the actual new case numbers are not consistently accurate over all counties and weeks. Second, the way counties are assigned demographic labels is subject to further debate. Third, county-week pairs with less than 100 new cases are removed. Fourth, an error metric that is customized based on what informs COVID-19 policy decisions may be more appropriate. Therefore, this case study should be viewed as a motivation for future studies rather than as conclusive evidence.

\section{Conclusion}\label{sec:con}

Despite the growing attention on machine learning fairness, most of the existing studies have focused on classification and ranking tasks. This paper, on the other hand, addresses the fairness testing problem in regression tasks. To this end, appropriate statistical hypothesis testing procedures are employed to determine whether the outputs from a decision system are fair towards different sensitive groups and to identify specific fairness issues and impacted groups. To illustrate the applicability and usefulness of the proposed methodology, a preliminary case study is conducted to investigate the relationship between the COVID-19 projections and the racial/ethnic demographics in the US. The findings revealed demographics-based disparities in the forecast errors.

Two areas of future work are anticipated. First, the proposed methodology tests for unfairness but does not necessarily quantify its extent. A line of future work may focus on designing and developing appropriate standardizable, generalizable, and interpretable fairness metrics to quantify unfairness in numeric prediction tasks. Second, the proposed fairness methodology can be employed to assess group fairness for regression tasks as part of algorithm and accountability audits of decision systems deployed in the real world. 

\section*{Acknowledgments}
This material is based upon work supported by the National Science Foundation under Grant CCF-2131504. Any opinions, findings, and conclusions or recommendations expressed in this material are those of the authors and do not necessarily reflect the views of the National Science Foundation.

\bibliographystyle{unsrt}
\bibliography{references}

\end{document}